%% file: root.tex
\documentclass[letterpaper, 10 pt, conference]{ieeeconf}  

\IEEEoverridecommandlockouts                              

\newcommand{\approach}{PersONAL~}

\title{\LARGE \bf PersONAL: Towards a Comprehensive Benchmark for Personalized Embodied Agents}

\author{Filippo Ziliotto$^{1,2}$, Jelin Raphael Akkara$^{1}$, Alessandro Daniele$^{2}$,
Lamberto Ballan$^{1}$, \\ Luciano Serafini$^{2}$, Tommaso Campari$^{2}$%
\thanks{$^{1}$University of Padova, Italy. 
{\tt\small jelinraphael.akkara@\ phd.unipd.it}, {\tt\small lamberto.ballan@unipd.it}}%
\thanks{$^{2}$Fondazione Bruno Kessler (FBK), Trento, Italy. 
{\tt\small \{fziliotto, daniele, serafini, tcampari\}@fbk.eu}}
}

\input{preamble}

\begin{document}

\maketitle
\thispagestyle{empty}
\pagestyle{empty}


\input{sec/0_abstract}
\input{sec/1_intro}
\input{sec/2_relworks}

\input{sec/3_task}
\input{sec/4_dataset}
\input{sec/5_experiments}

\input{sec/6_conclusions}





\section*{ACKNOWLEDGMENTS}
We thank the University of Padua, Department of Mathematics “Tullio Levi-Civita”, for providing the computational resources. TC and LS were supported by the PNRR project Future AI Research (FAIR - PE00000013) under the NRRP MUR program, funded by NextGenerationEU.


\bibliographystyle{IEEEtran} 
\bibliography{IEEEfull}




\end{document}

%% file: preamble.tex
\usepackage{graphicx}     
\usepackage{booktabs}     
\usepackage{pifont}       
\usepackage[accsupp]{axessibility} 
\usepackage{array,multirow}
\usepackage{colortbl}
\usepackage{wrapfig}
\usepackage{comment}
\usepackage{amsmath,amssymb}
\usepackage{algorithm}
\usepackage[noend]{algpseudocode}
\usepackage{times}        
\usepackage[table]{xcolor}
\usepackage{tabularx}
\usepackage{cite}
\usepackage{paralist}

\usepackage[colorlinks=true,
            linkcolor=black,
            urlcolor=red,
            citecolor=green]{hyperref}

\newcolumntype{C}{>{\centering\arraybackslash}X}

\newcommand{\cmark}{\ding{51}}%
\newcommand{\xmark}{\ding{55}}%

%% file: sec/0_abstract.tex
\begin{abstract}

Recent advances in Embodied AI have enabled agents to perform increasingly complex tasks and adapt to diverse environments. However, deploying such agents in realistic human-centered scenarios, such as domestic households, remains challenging, particularly due to the difficulty of modeling individual human preferences and behaviors.
In this work, we introduce \approach~(\emph{PERS}onalized \emph{O}bject \emph{N}avigation \emph{A}nd \emph{L}ocalization), a comprehensive benchmark designed to study personalization in Embodied AI.
Agents must identify, retrieve, and navigate to objects associated with specific users, responding to natural-language queries such as \emph{find Lily’s backpack.}
\approach comprises over 2,000 high-quality episodes across 30+ photorealistic homes from the HM3D dataset. Each episode includes a natural-language scene description with explicit associations between objects and their owners, requiring agents to reason over user-specific semantics.

The benchmark supports two evaluation modes: (1) active navigation in unseen environments, and (2) object grounding in previously mapped scenes.
Experiments with state-of-the-art baselines reveal a substantial gap to human performance, highlighting the need for embodied agents capable of perceiving, reasoning, and memorizing over personalized information; paving the way towards real-world assistive robot.
Code and dataset available at: \href{https://github.com/ZiliottoFilippoDev/PersONAL}{github.io/PersONAL}
\end{abstract}

\begin{keywords}
Embodied AI, Personalized Agents, Benchmarks
\end{keywords}

%% file: sec/1_intro.tex
\section{INTRODUCTION}
\label{sec:intro}



In recent years, Embodied AI has significantly advanced, enabling agents to perform complex tasks and interact more naturally with their environments. Modern methods combine end-to-end training with zero-shot capabilities powered by large language models (LLMs), allowing agents to answer dynamically to user input~\cite{yang20253d, liang2023code, song2023llm, ziliotto2025tango, li2024tina, wang2024vlm, kim2024survey}. 
Yet, their application to user-centric scenarios, where agents must interpret local, implicit information not directly encoded in pretrained models, such as object ownership, remains largely unexplored. 
Bridging this gap is key to deploying embodied agents in real-world environments like homes or offices.
While personalized vision-language models (VLMs)~\cite{pham2025plvm, alaluf2024myvlm, pham2024personalized} have been developed for user-specific visual grounding, they are typically limited to static, image-based contexts. 
In contrast, embodied agents must operate in complex physical settings, reasoning and acting over time. 

Recently, a few works have begun to explore personalization in embodied scenarios~\cite{barsellotti2024personalized, dai2024think}, but the field remains in its early stages and these works mainly focus on guiding agents using image-based queries or continuous human-robot interaction, which limits scalability and real-world applicability. 

\begin{figure}[!t]
    \centering
    \includegraphics[width=1.\columnwidth]{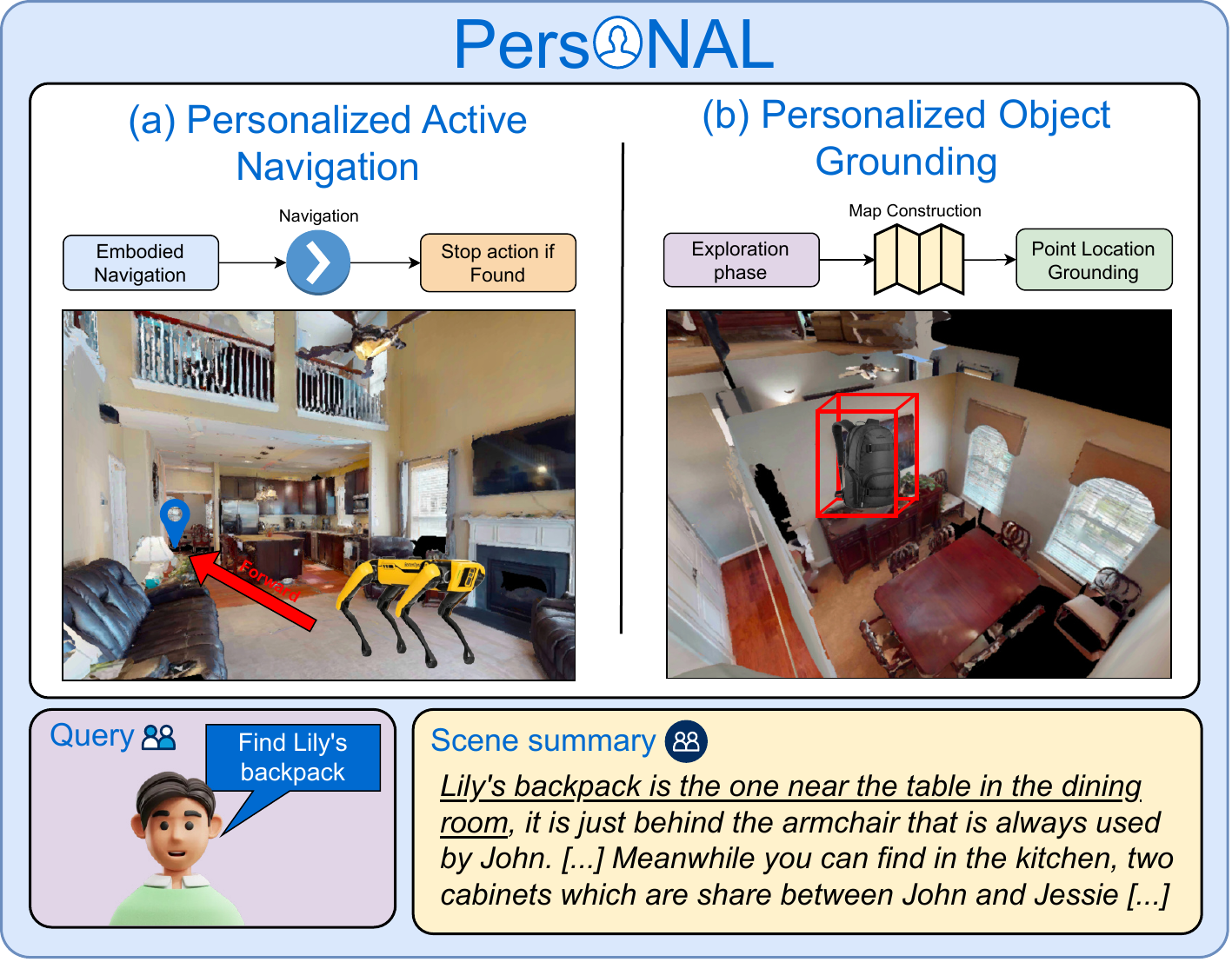}
    \caption{We introduce \textbf{\approach}, a comprehensive benchmark to evaluate Embodied AI agents in the context of user centric tasks. \approach supports two evaluation modes: (1) active navigation in unseen environments, and (2) object grounding in previously mapped scenes. }
    \label{fig:teaser}
\end{figure}



To address this gap, we introduce \approach, the first embodied AI benchmark for personalized, user-centric navigation and personalized object grounding (Figure~\ref{fig:teaser}). 
Agents must interpret user-specific queries and either navigate to or retrieve the location of objects associated with particular individuals (e.g., “Navigate to Carl’s backpack”). 
Each episode includes a textual scene description specifying object attributes and ownership (e.g., “the upper kitchen cabinet belongs to Linda”), followed by a personalized query.
Unlike prior work, we also define a grounding task which acts as an embodied memory challenge, requiring agents to recall and localize targets using internal maps.


\approach supports two evaluation modes: \emph{(i)} Personalized Active Navigation (PAN), the agent must navigate an unknown environment to find the target object, and \emph{(ii)} Personalized Object Grounding (POG), where the goal is to localize objects within a pre-mapped environment.
We release a dataset of over 2,000 curated evaluation episodes across three difficulty levels (easy, medium, hard), capturing increasing complexity in human–object associations
We evaluate the proposed benchmark with several zero-shot, state-of-the-art navigation baselines and introduce a simple zero-shot method for the grounding task, demonstrating in both settings that a substantial performance gap persists between current Embodied AI agents and human capabilities.

In summary, our main contributions are:
\begin{itemize}
\item We present \approach, a comprehensive Embodied AI benchmark specifically designed to evaluate embodied personalization, incorporating user-centric queries and object–ownership semantics.
\item We release a dataset of 2,000 high-quality episodes sampled from over 30 realistic household environments, divided into three difficulty levels (easy, medium, and hard).
\item We provide empirical analyses with state-of-the-art zero-shot baselines, showcasing limitations and possible future research directions toward human-level personalized navigation in Embodied AI.
\end{itemize}

We argue that personalization cannot be achieved through repeated LLM-based interactions alone. 
Instead, agents must maintain and recall user-specific preferences internally—ideally through long-term memory and continual learning—an ability crucial in complex households settings where ownership is stable but scenes continually evolve.



%% file: sec/2_relworks.tex
\section{RELATED WORKS}
\label{sec:rel_works}

Research in embodied visual navigation has accelerated with the emergence of large-scale photorealistic simulators and datasets~\cite{savva2019habitat, kolve2017ai2thor, ramakrishnan2021habitat, chang2017matterport}, enabling a range of navigation tasks including point-goal, object-goal, image-goal, and language-guided navigation~\cite{batra2020objectnav, krantz2022instance, anderson2018vision, ramakrishnan2021habitat, wani2020multion}. 
Therefore, we review prior work in embodied AI, with a particular focus on navigation tasks, and discuss related research on personalization within computer vision.

\subsection{Embodied Navigation}

Progress in embodied navigation has been driven by advanced simulation platforms and rich 3D datasets. AI2-THOR~\cite{kolve2017ai2thor} and Habitat~\cite{savva2019habitat} provide efficient, photorealistic environments supporting large-scale training and evaluation of navigation, manipulation, and visual QA tasks. Datasets like Matterport3D, Gibson, and the larger, more diverse HM3D~\cite{chang2017matterport, batra2020objectnav} enable agents to learn and generalize in realistic settings, with HM3D offering over $112,000,m^2$ of navigable space. These resources support reproducible research in ObjectNav, ImageNav, and grounding tasks~\cite{chaplot2020object, krantz2022instance, krantz2023navigating, wani2020multion}.

Recent benchmarks have further increased realism and complexity, for example, the Open-Vocabulary ObjectNav challenge~\cite{yokoyama2024hm3d} uses HM3D with 15k object instances across 379 categories and free-form text goals. 
GOAT-Bench~\cite{khanna2024goat} introduces multimodal target sequences, while zero-shot navigation has been explored by ZSON~\cite{majumdar2022zson} and ORION~\cite{dai2024think}.
On the other hand, the FindingDory Benchmark~\cite{yadav2025findingdory} evaluates the long-term memory capabilities of embodied agents, requiring tasks such as``navigate to the first room you visited'' after exploration. 
However, all these works lack the focus on personalization capabilities that embodied agents require to function effectively in real-world households or office environments (e.g., ``find Lisa’s phone and bring it to her'').

\subsection{User-centric AI Agents}

Prior work on personalized vision-language models (VLMs)~\cite{pham2025plvm, alaluf2024myvlm, pham2024personalized} has focused on user-specific visual grounding in static, image-based settings. 
However, personalization in embodied AI—where agents must interpret and act on user queries in dynamic physical environments—remains largely unexplored. 
Our work addresses this gap by extending personalization to real-world scenarios.

Standard Embodied AI typically treats targets as generic instances (e.g., any chair), without distinguishing ownership~\cite{batra2020objectnav, wani2020multion, khanna2024goat}. 
Recent work has begun to address this by introducing Personalized Instance-based Navigation~\cite{barsellotti2024personalized}, where agents must locate specific user-owned objects among multiple similar items, inputted as images.

Other works have incorporated personalization in object-goal navigation~\cite{dai2024think}, but often these benchmarks are not always rigorously defined and depend on continuous user interaction through LLM-based API calls, which is impractical for real-world deployment. 
Furthermore, such queries do not account for scenarios involving multiple objects owned by different individuals.
Instead, we advocate for approaches like~\cite{chang2023goat}, where agents can remember user-centric habits and learn about the environment over time without external dependencies.

User-centric embodied AI requires adapting agent behavior to individual preferences, such as caution or speed. While recent work~\cite{hwang2024promptable} explores multi-objective reinforcement learning to adapt agents without retraining, personalization remains underexplored and is still in its early stages, with ongoing efforts to extend it across modalities and scenarios toward more human-aligned agents.
In this work, we present \approach, a robust benchmark for Embodied AI aimed at advancing research in this area.

%% file: sec/3_task.tex
\section{Benchmark}
\label{sec:benchmark}

\begin{figure}[t]
    \centering
    \includegraphics[width=1.\columnwidth]{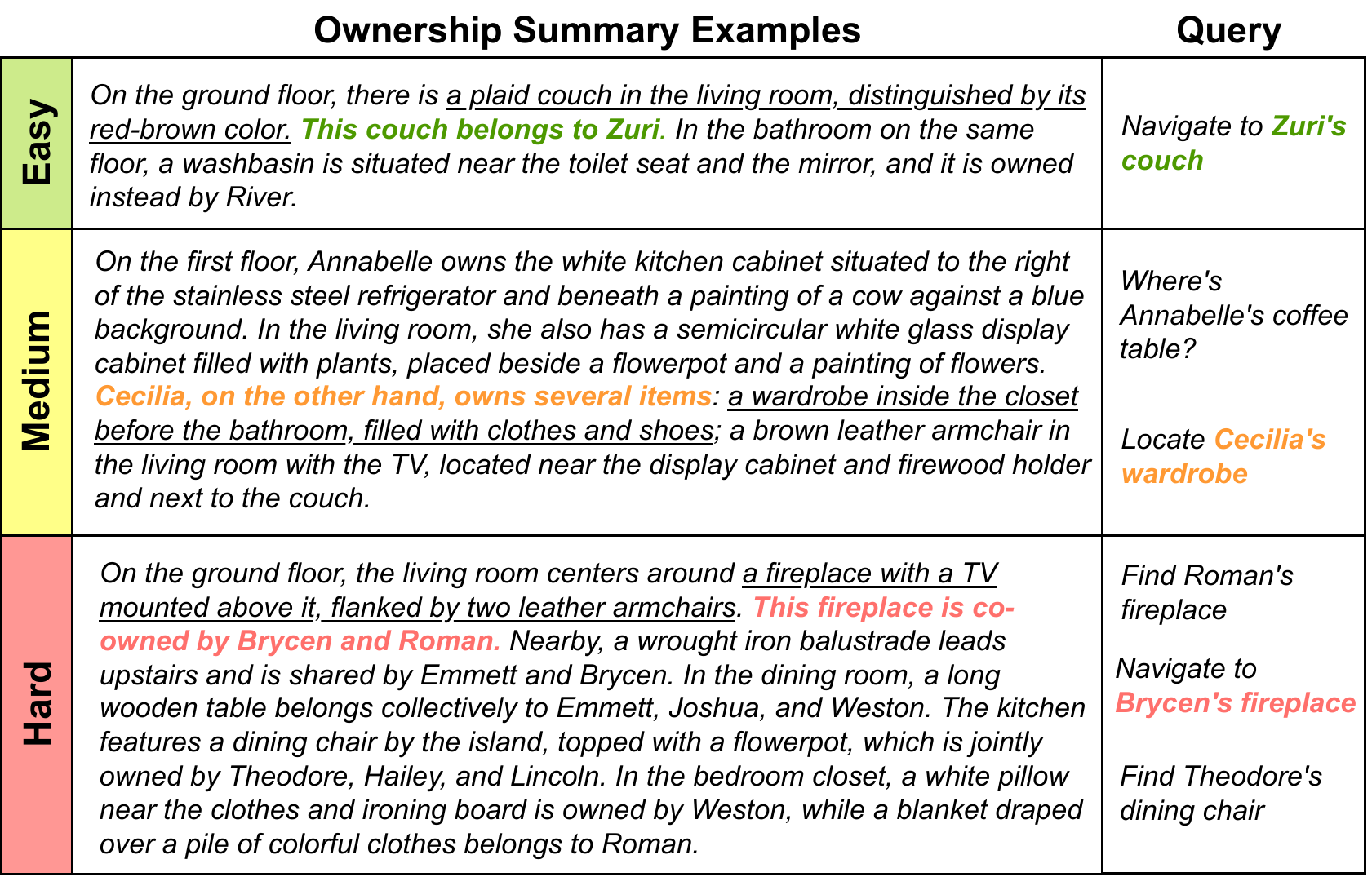}
    \caption{\textbf{Example episodes from \approach.} The agent receives, as input, a personalized query along with a scene description specifying ownership details. Both navigation and grounding tasks are conditioned on this information.}
    \label{fig:split_example}
\end{figure}

In this section, we introduce the proposed \approach benchmark and its associated tasks. 
The benchmark is structured into two main components: (1) Personalized Active Navigation (PAN), where the agent must navigate to a user-specified object without any prior knowledge of the environment; and (2) Personalized Object Grounding (POG), where the agent can pre-explore the scene to build a representation of the environment (e.g., a map) to assist in localizing the target afterwards.

In both settings, the agent receives a detailed scene description specifying object ownership (e.g., ``in the kitchen, the cabinet on the left belongs to Lisa, while the one on the right belongs to James. Meanwhile, the picture above the bed belongs to ...''), as well as a user query such as ``Find Lisa's cabinet.'' In the POG task, the query and the scene description are provided \emph{after} the pre-exploration phase.
The agent must correctly interpret both the scene description and the query to solve the task, as shown in Figure~\ref{fig:split_example}.
In contrast to previous work~\cite{barsellotti2024personalized}, our approach provides the object reference exclusively in textual form, mirroring how humans typically communicate ownership or describe personal items (e.g., ``my bag is the one with blue and white stripes''). 
We deliberately avoid relying on image-based cues, which are uncommon and often impractical in real-world scenarios, since it requires collecting images for every single object belonging to a specific user.

\subsection{Personalized Active Navigation (PAN)}

In the Personalized Active Navigation (PAN) setting, the agent is given a budget of 500 steps to reach the target location from its initial position. 
The agent must rely solely on the given personalized information, without any further user interaction during the episode. 
This constraint reflects realistic use cases—after receiving an initial description, a robot should autonomously identify and localize the target object without continuous user input confirmation.

The task settings follow standard ObjectGoal Navigation conventions~\cite{batra2020objectnav}.
Within the Habitat framework, we set the agent’s camera height to 0.88 meters for both RGB and Depth sensors, with a field of view (FOV) of 79°, imitating the embodiment of a LocoBot robot. 
The action space consists of MOVE\_FORWARD (0.25 m), TURN\_LEFT and TURN\_RIGHT (30° each), LOOK\_UP and LOOK\_DOWN (30° each), and the STOP action. 
An episode is considered successful if, after selecting the STOP action, the agent lies within 1.0 meters from the target object.

\subsection{Personalized Object Grounding (POG)}

In the Query Grounding task, the agent is first allowed to pre-explore the environment with a budget of at most 2,500 steps, during which it can construct a spatial representation of the environment. 
We note that the robot configuration in the pre-exploration phase is identical to that used in the PAN task.
We place no restrictions on the map representation: it can be a compressed 2D top-down view derived from the point cloud, a full 3D reconstruction, or even a video sequence with frame-by-frame updates. 
We developed a zero-shot approach to address the first representation, while the latter options are left for future work.
Following the exploration phase, the agent is presented with an input query and is required to localize the corresponding target object by outputting its coordinates on the given map.


%% file: sec/4_dataset.tex
\section{Dataset}
\label{sec:dataset}

\begin{figure}[t]
    \begin{center}
    \includegraphics[width=\columnwidth]{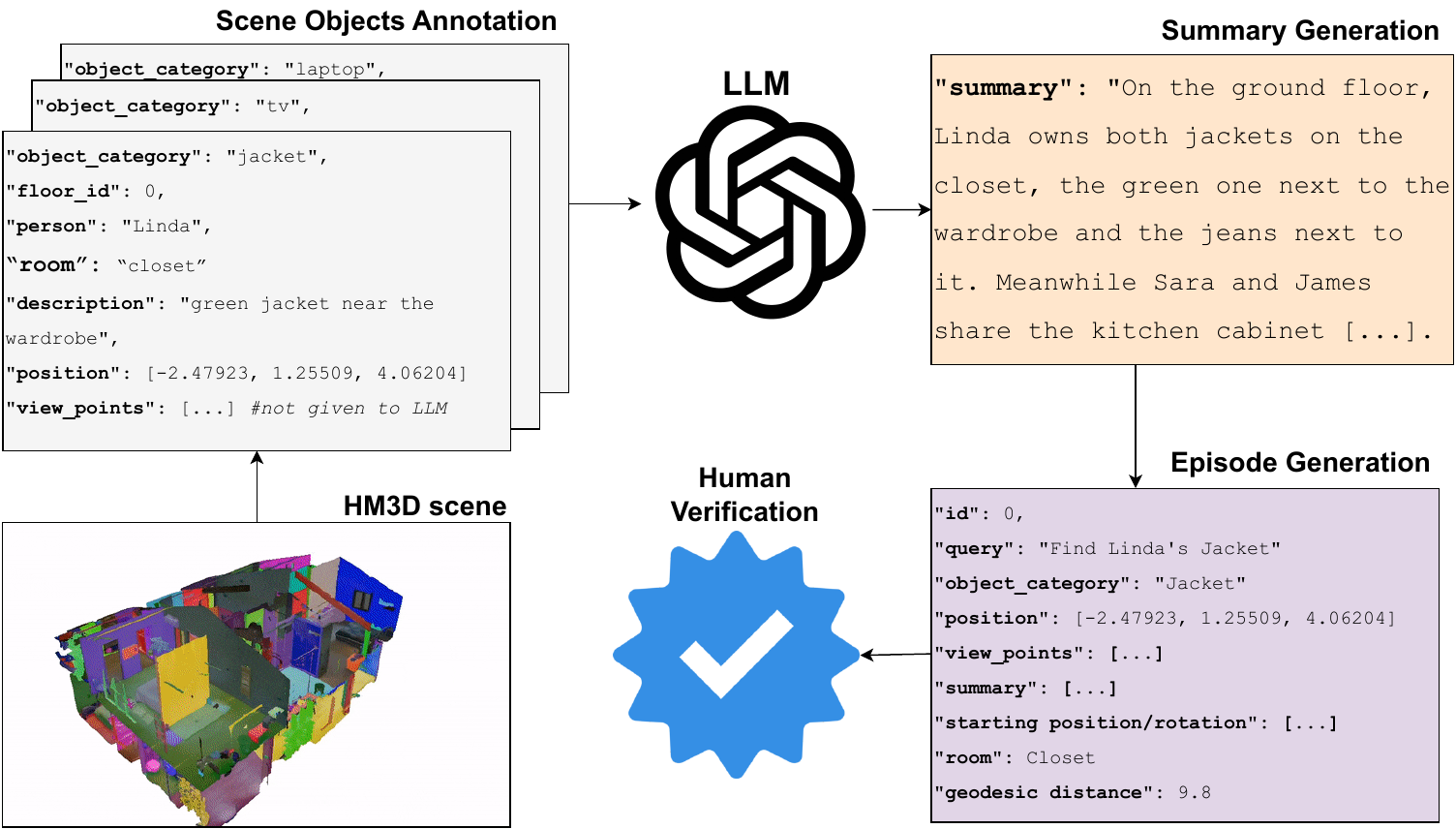}
    \caption{\textbf{\approach Dataset generation}. We selected a substantial subset of episodes to manually assess the quality of the generated episodes.}
    \label{fig:dataset_generation}
    \end{center}
\end{figure}

\begin{figure}[t]
    \centering
    \includegraphics[width=1.\columnwidth]{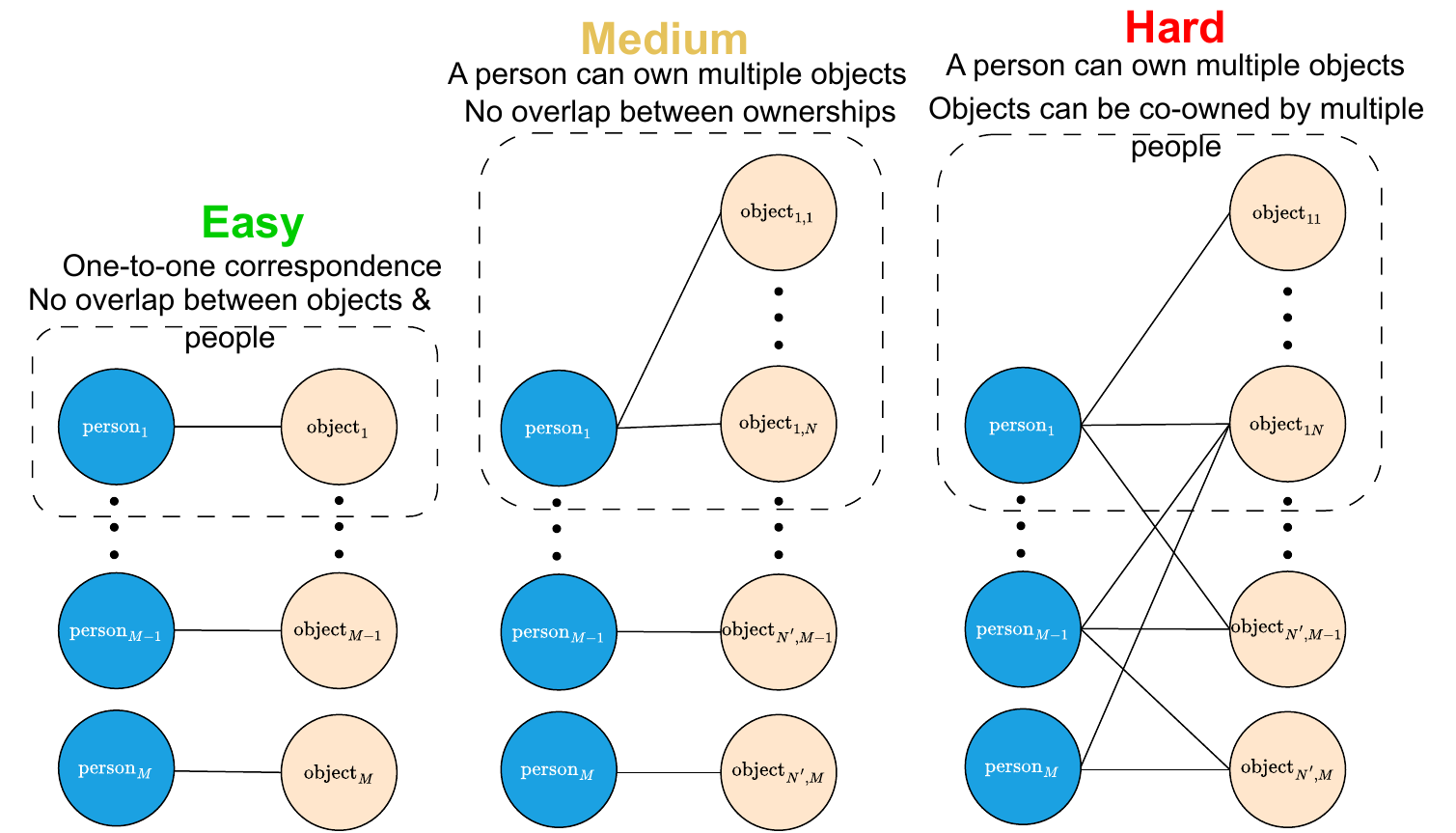} 
    \caption{\textbf{\approach Dataset splits}. Difficulty levels are represented as bipartite graphs between objects and owners.}
    \label{fig:dataset_splits}
\end{figure}

Here we describe the composition and generation process of the \approach dataset.
We curated over 2,000 episodes designed to evaluate agents’ ability to interpret user queries in the context of personalized human–object ownership information.
To ensure the benchmark remains challenging and broadly useful, the dataset is divided into three difficulty levels: ``easy'', ``medium'', and ``hard'' (see Figure~\ref{fig:dataset_splits}).
The core idea is that personalization can, in principle, be addressed by storing user–object associations in an explicit memory or database. 
Or similarly, using model adaptation methods like adapters~\cite{rebuffi2017learning, hu2022lora}.
However, naive memorization does not scale, and the three difficulty levels are specifically designed to encourage methods that achieve efficient personalization.

\approach builds on the work of GOAT-Bench dataset~\cite{khanna2024goat}, which provides textual descriptions for objects within HM3D environments.
In particular, it comprises over 140 object categories distributed across three validation splits: ``seen,'' ``seen\_synonyms,'' and ``unseen.''
However, during the dataset generation, we observed that a substantial fraction of these original descriptions were either non-informative or factually incorrect; specifically, we found that approximately half of the provided captions suffered from these issues. 

\noindent\textbf{Caption Refinement.} To address this, we developed a dedicated annotation tool that enabled manual refinement of object captions. 
This process focused on producing more realistic, descriptive, and spatially grounded annotations (e.g., ``the red backpack under the table''), in line with the overall goal of \approach: to introduce realistic and actionable human-centric descriptions into embodied navigation tasks.
We manually refined the captions for over 3,000 individual objects, discarding non-informative or irrelevant entries to ensure high-quality and meaningful annotations throughout the dataset. 
As shown in Table~\ref{tab:goat_comparison}, our dataset exhibits greater lexical diversity and less repetition, with a more balanced distribution of word usage compared to GOAT-Bench. 
In particular, Entropy increased from 6.92 to 7.35 bits (a +6\% change), corresponding to a ~35\% increase in the effective diversity of word usage (line 3).
Moreover, the average number of words per description decreased, along with the standard deviation, indicating that the descriptions are more coherent and less redundant (line 1). 
In contrast, GOAT-Bench often contains descriptions that are unnecessarily long or repetitive.
Unlike prior works, we also labeled each object with its room and floor in an open-set manner (e.g., ``the master bedroom'', ``the child’s bed''), yielding more realistic and context-rich scene descriptions in the second phase of the dataset generation.
Specific differences with respect to other benchmarks are shown in Table~\ref{tab:benchmark_comparison}.


\setlength{\tabcolsep}{5pt}
\begin{table}[t]
\centering
\scriptsize
\begin{tabular}{lccc}
\toprule
\cellcolor{gray!10}\textbf{Metric} & \cellcolor{gray!10}\textbf{GOAT-Bench}~\cite{khanna2024goat} & \cellcolor{gray!10}\textbf{\approach (Ours)} & \cellcolor{gray!10}\textbf{Gain} \\
\midrule
Words / Description $(\downarrow)$  & $20.65 \pm 13.10$      & $16.28 \pm 6.06$ & -21\% \\
Vocabulary size $(\uparrow)$        & 1344  & 1564  & +16\% \\
Type--Token Ratio $(\uparrow)$      & 0.051 & 0.069 & +36\% \\
Shannon entropy (bits) $(\uparrow)$ & 6.92  & 7.35  & +6\% \\
Faulty Captions $(\downarrow)$      & 40\%  & 0\%   & -- \\
\bottomrule
\end{tabular}
\caption{\textbf{Description quality comparison between GOAT-Bench and \approach.} Gains are relative improvements of \approach over GOAT-Bench.}
\label{tab:goat_comparison}
\end{table}

\noindent\textbf{Episode Personalization.} In the second phase of dataset generation, we assigned object ownership to specific individuals for each episode, thereby establishing personalized ownership relations. 
Ownership assignments are modeled as a bipartite graph, with constraints applied according to the chosen split difficulty (see Figure~\ref{fig:dataset_splits}). 
In the ``easy'' split each owner possesses at most one object and no overlap between object assignments, having a one-to-one correspondence.
The ``medium'' split allows each owner to possess multiple objects, but each object can still belong to only one person. 
In the ``hard'' split, all constraints are relaxed: owners may possess multiple objects, and objects may be shared among multiple people.

In mathematical terms, let $A \in \{0,1\}^{M \times N}$ be the ownership matrix, where $M$ is the number of people and $N$ the number of objects. Define
\[
r_i = \sum_{j=1}^N A_{ij}, \qquad c_j = \sum_{i=1}^M A_{ij},
\]
where $r_i$ is the number of objects owned by person $i$, and $c_j$ is the number of people owning object $j$. The constraints for each difficulty split are:
\[
\small
\begin{cases}
\text{Easy:}   & \forall i:\ r_i = 1;\ \forall j:\ c_j = 1;\ M = N.\\
\text{Medium:} & \forall i:\ r_i \geq 1;\ \forall j:\ c_j = 1;\ \exists\, i:\ r_i > 1.\\
\text{Hard:}   & \forall i:\ r_i \geq 1;\ \forall j:\ c_j \geq 1;\ \exists\, i:\ r_i > 1;\ \exists\, j:\ c_j > 1.
\end{cases}
\]

Finally, to generate each episode, we input both the ownership graph and the refined object descriptions into a large language model (GPT-4.1 in our case), prompting it to produce a comprehensive scene description that explicitly encodes the ownership associations (see Figure~\ref{fig:dataset_generation}). 
Each generated description is then paired with a corresponding query targeting a specific ownership relation. 
To ensure data quality, we manually reviewed a subset of 300 episodes after generation to assess the accuracy and correctness of the resulting ``scene summaries'', with more than 90\% of the episodes correctly labelled.
Moreover, in order to enhance variability across episodes, some scene descriptions include multiple instances of the same object category, which may be owned by a single individual or by multiple people, e.g. ``Find one of Julia's beds''.
The main dataset statistics are shown in Table~\ref{tab:network-metrics}.

\begin{table}[t]
  \scriptsize
  \setlength{\tabcolsep}{5pt}
  \centering
  \begin{tabularx}{\columnwidth}{lCCCCCCCCCCCC}
    \toprule
    \textbf{Feature} &
    \rotatebox{90}{\cellcolor{gray!10}\textbf{\approach (Ours)}} &
    \rotatebox{90}{PInNED~\cite{barsellotti2024personalized}} &
    \rotatebox{90}{GOAT-Bench~\cite{khanna2024goat}} &
    \rotatebox{90}{ZIPON~\cite{dai2024think}} &
    \rotatebox{90}{InstImageNav~\cite{krantz2022instance}} &
    \rotatebox{90}{ProcTHOR~\cite{deitke2022procthor}} &
    \rotatebox{90}{ION~\cite{li2021ion}} &
    \rotatebox{90}{ZSON~\cite{majumdar2022zson}} &
    \rotatebox{90}{AI2-THOR~\cite{kolve2017ai2thor}} &
    \rotatebox{90}{Gibson~\cite{gibson}} &
    \rotatebox{90}{Robo-THOR~\cite{deitke2020robothor}} &
    \rotatebox{90}{MultiON~\cite{wani2020multion}} \\
    \midrule
    Photo-Realistic Scenes & \cellcolor{gray!10}\cmark & \cmark & \cmark & \cmark & \cmark & \xmark & \xmark & \cmark & \xmark & \cmark & \xmark & \cmark \\
    Object Caption         & \cellcolor{gray!10}\cmark & \cmark & \cmark & \cmark & \xmark & \xmark & \cmark & \cmark & \xmark & \xmark & \xmark & \cmark \\
    Room \& Floor Annots.  & \cellcolor{gray!10}\cmark & \xmark & \xmark & \xmark & \xmark & \cmark & \xmark & \xmark & \cmark & \cmark & \xmark & \cmark \\
    Scene Description      & \cellcolor{gray!10}\cmark & \xmark & \xmark & \xmark & \xmark & \xmark & \xmark & \xmark & \xmark & \xmark & \xmark & \xmark \\
    Object Ownership       & \cellcolor{gray!10}\cmark & \cmark & \xmark & \cmark & \xmark & \xmark & \xmark & \xmark & \xmark & \xmark & \xmark & \xmark \\
   Navigation Task         & \cellcolor{gray!10}\cmark & \cmark & \cmark & \cmark & \cmark & \cmark & \cmark & \cmark & \cmark & \cmark & \cmark & \cmark \\
    Grounding Task         & \cellcolor{gray!10}\cmark & \xmark & \xmark & \xmark & \xmark & \xmark & \xmark & \xmark & \xmark & \xmark & \xmark & \xmark \\
    \bottomrule
  \end{tabularx}
  \caption{\textbf{Comparison of \approach with existing embodied datasets}. Features are shown as rows, datasets as columns (rotated for compactness).}
  \label{tab:benchmark_comparison}
\end{table}

\setlength{\tabcolsep}{6.5pt}
\begin{table}[t]
  \centering
  \small
  \begin{tabular}{lcccc}
    \toprule
     & \multicolumn{3}{c}{\textbf{Split}} &  \\
    \cmidrule(lr){2-4}
    \textbf{\approach Statistics} & \textbf{Easy} & \textbf{Medium} & \textbf{Hard} &  \textbf{Total}\\
    \midrule
    \multicolumn{5}{l}{\cellcolor{gray!10}\emph{Episode Entities}} \\
    \midrule
    N° People              & 4.52 & 4.42 & 6.27 & 5.07 \\
    N° Objects             & 4.52 & 6.09 & 6.42 & 5.68 \\
    N° Shared Objects      & 0.00 & 0.00 & 4.55 & 1.52 \\
    N° Object Categories   & 51   & 83   & 119  & 119 \\
    Avg.\ Summary Length   & 67   & 117  & 148  & 110 \\
    \midrule
    \multicolumn{5}{l}{\cellcolor{gray!10}\emph{Ownership Graph}} \\
    \midrule
    N° Edges               & 4.52 & 6.09 & 16.42 & 9.01 \\
    Avg.\ Degree per Person  & 1.00 & 1.43 & 2.61 & 1.68 \\
    Avg.\ Degree per Object  & 1.00 & 1.00 & 2.62 & 1.54 \\
    Density                & 0.23 & 0.24 & 0.42 & 0.30 \\
    Overlap Ratio          & 0.00 & 0.00 & 0.73 & 0.24 \\
    \midrule
    \multicolumn{5}{l}{\cellcolor{gray!10}\emph{Navigation}} \\
    \midrule
    N° Episodes            & 600  & 700  & 720  & 2020 \\
    Avg.\ Geodesic Distance   & 3.12 & 5.23 & 7.56 & 5.30 \\
    Avg.\ Euclidean Distance  & 3.40 & 4.80 & 7.09 & 5.10 \\
    \bottomrule
  \end{tabular}
  \caption{\textbf{Dataset metrics by task difficulty}. It comprises graph structure variables and navigation settings entities.}
  \label{tab:network-metrics}
\end{table}

Regarding the dataset construction for the active navigation pipeline (PAN) we generated both starting positions and valid viewpoints for each episode. 
However, GOAT-Bench~\cite{khanna2024goat} scenes do not provide annotated viewpoints for all the considered objects. 
To generate them, we sample candidate positions and project a line from each to the object’s boundary points (i.e. its center position). 
A viewpoint is retained only if the line does not intersect walls or is not heavily occluded by other objects, ensuring clear visibility of targets. 
We then discarded all the viewpoints distanced more than 1m to the target object center. 
In specific cases, we relax the 1m threshold to 1.5m, since HM3D provides only the center position of each object and not its dimensions. For larger objects, such as pictures or couches, the standard 1m constraint may exclude valid points; relaxing the threshold ensures a fair evaluation in these scenarios.

%% file: sec/5_experiments.tex
\section{EXPERIMENTS}
\label{sec:experiments}

In this section, we describe the baseline approaches evaluated on the proposed \approach dataset.
For the active navigation task, we consider established zero-shot ObjectNav methods~\cite{wang2024vlm, busch2024one} that combine end-to-end reinforcement learning policies with vision–language models (VLMs) or related mechanisms to guide exploration toward promising frontiers.
For the grounding task, we introduce a simple baseline that leverages frontier exploration policies in combination with a memory module constructed from VLM embedding vectors.


\subsection{PAN Baselines}
\noindent\textbf{Random}. As a sanity check, we first implemented a Random baseline, where the agent selects actions uniformly at random from all available frontiers and terminates after 500 steps, in order to avoid biases due to varying initial distances.

\noindent\textbf{Human}. To establish an upper bound on performance, we also evaluated human participants on the benchmark. 
Specifically, we sampled 50 episodes from each difficulty split and measured the performance of five human subjects, reporting the average as the final baseline. At each step, participants were provided with the current RGB image and the four possible agent actions.

\noindent\textbf{VLFM}. We employed a state-of-the-art zero-shot navigation approach that constructs occupancy maps from depth observations to detect exploration frontiers, while using RGB observations and a pre-trained vision-language model to generate a language-grounded value map. VLFM uses this value map to prioritize frontiers most likely to contain the queried object category.
For a fair comparison, in order to define the target category for each query, we use an LLM to extract the most appropriate instance (e.g., for the query ``where's Linda's clothes,'' the LLM identifies the related association ``green jacket in the closet'' as the target from the scene description). It then localizes the objects using open-set detectors~\cite{minderer2024scaling}.
To detect the objects extracted by the LLM, we used an open-set object detector, namely Owl-ViT2~\cite{minderer2024scaling}. 
We also experimented with other detectors (e.g., GroundingDINO~\cite{liu2024grounding}), but they yielded lower overall performance.

\noindent\textbf{OneMap}. Similarly to VLFM, this zero-shot method constructs an enhanced and reusable open-vocabulary feature map for real-time object search, using a probabilistic semantic map update to reduce errors in feature extraction. 
This approach incorporates semantic uncertainty, enabling more informed exploration and higher path efficiency.
As before we leverage an LLM to fetch the open-set target object category.

\setlength{\tabcolsep}{5pt}
\begin{table*}[!ht]
\centering
\small
\begin{tabular}{l *{12}{c}}
\toprule
\multicolumn{1}{c}{} &
\multicolumn{3}{c}{\textbf{Easy}} &
\multicolumn{3}{c}{\textbf{Medium}} &
\multicolumn{3}{c}{\textbf{Hard}} &
\multicolumn{3}{c}{\textbf{Total}} \\                                                                
\cmidrule(lr){2-4}\cmidrule(lr){5-7}\cmidrule(lr){8-10}\cmidrule(lr){11-13}
\textbf{Method} & SR$\tiny\uparrow$ & SPL$\tiny\uparrow$ & DTG$\tiny\downarrow$ & SR$\tiny\uparrow$ & SPL$\tiny\uparrow$ & DTG$\tiny\downarrow$ & SR$\tiny\uparrow$ & SPL$\tiny\uparrow$ & DTG$\tiny\downarrow$ & SR$\tiny\uparrow$ & SPL$\tiny\uparrow$ & DTG$\tiny\downarrow$ \\
\midrule
\cellcolor{gray!10} Human                           
& \cellcolor{gray!10} 70.21 & \cellcolor{gray!10} 49.20 & \cellcolor{gray!10} 1.26
& \cellcolor{gray!10} 65.22 & \cellcolor{gray!10} 40.05 & \cellcolor{gray!10} 1.40
& \cellcolor{gray!10} 60.13 & \cellcolor{gray!10} 35.31 & \cellcolor{gray!10} 1.32
& \cellcolor{gray!10} 63.54 & \cellcolor{gray!10} 41.34 & \cellcolor{gray!10} 1.48 \\

\cellcolor{gray!10} Human \textit{w. env. memory}$^\dagger$  
& \cellcolor{gray!10} 89.00 & \cellcolor{gray!10} 58.12 & \cellcolor{gray!10} 0.84
& \cellcolor{gray!10} 86.00 & \cellcolor{gray!10} 55.00 & \cellcolor{gray!10} 1.10
& \cellcolor{gray!10} 82.00 & \cellcolor{gray!10} 53.20 & \cellcolor{gray!10} 0.98
& \cellcolor{gray!10} 86.00 & \cellcolor{gray!10} 52.18 & \cellcolor{gray!10} 1.03 \\
\midrule
Random                          &   0.01   &  0.00    &   6.02   &   0.05   &  0.00    &  6.98    &  0.01   &  0.00    &   0.00   &  0.00    &   0.00   &   8.20   \\
VLFM~\cite{yokoyama2024vlfm}  &  12.50    & 6.00     &  5.21    &  10.49    &  4.00    &  5.18    &  7.82    & 1.00     &   5.03   &   9.88   &   6.00   & 5.03 \\
OneMap~\cite{busch2024one}  &  \textbf{38.17}    &  \textbf{29.00}    &  \textbf{3.04}    &  \textbf{28.61}    &  \textbf{21.00}    &  \textbf{3.75}    & \textbf{20.00}     &  \textbf{16.00}    &   \textbf{4.68}   &  \textbf{28.93}  & \textbf{22.00} & \textbf{3.82}  \\
\bottomrule
\end{tabular}
\caption{\textbf{PAN Results.} $^\dagger$Evaluations conducted with prior environment knowledge, as the map was shown before the episode start.}
\label{tab:pon_results}
\end{table*}

\subsection{POG Baselines}

In this setting, all baseline maps are constructed uniformly.
The agent first explores the environment using a frontier-based policy~\cite{yamauchi1998frontier}, generating a 3D point cloud from depth observations, which is projected into a 2D top-down map of navigable and non-navigable areas. This spatial map is discretized into 1×1 m grid cells, each storing a semantic embedding extracted from the agent’s egocentric RGB observations via a frozen vision–language model (BLIP-2~\cite{li2023blip}). The resulting feature map has dimensions 50×50×768, providing a compact spatial memory that supports efficient retrieval for grounding. As noted earlier, we do not impose constraints on the mapping itself, since the task is solely to predict the correct real-world coordinates of the queried target object.

\noindent\textbf{Random}. To control for potential biases, we evaluated two random baselines: (i) an agent that selects a random point in the 2D top-down map, and (ii) an agent that always predicts the map center. The latter is included to account for cases where small maps might make the center a disproportionately strong predictor of the ground-truth object location, which we aim to avoid.

\noindent\textbf{Human}. The human baseline in this setting is established by providing participants with visualizations of the GLB files of the HM3D scenes. 
Using an automated script, we record the predicted location when the user selects the target area with a mouse click, and then compute the relevant evaluation metrics accordingly.

\noindent\textbf{Query Scoring (QS)}.
As a naive zero-shot baseline, we scored the query embeddings (encoded with a VLM encoder) directly against the feature map vectors. The same encoder was used on both sides to ensure embedding alignment. The predicted location was taken as the index of the maximum similarity in the output map and compared to the ground-truth object position. Although this approach ignores the scene description, it serves to confirm that simple query matching alone cannot solve the dataset and that no shortcut exists.

\begin{figure}[ht]
    \centering
    \includegraphics[width=1.\columnwidth]{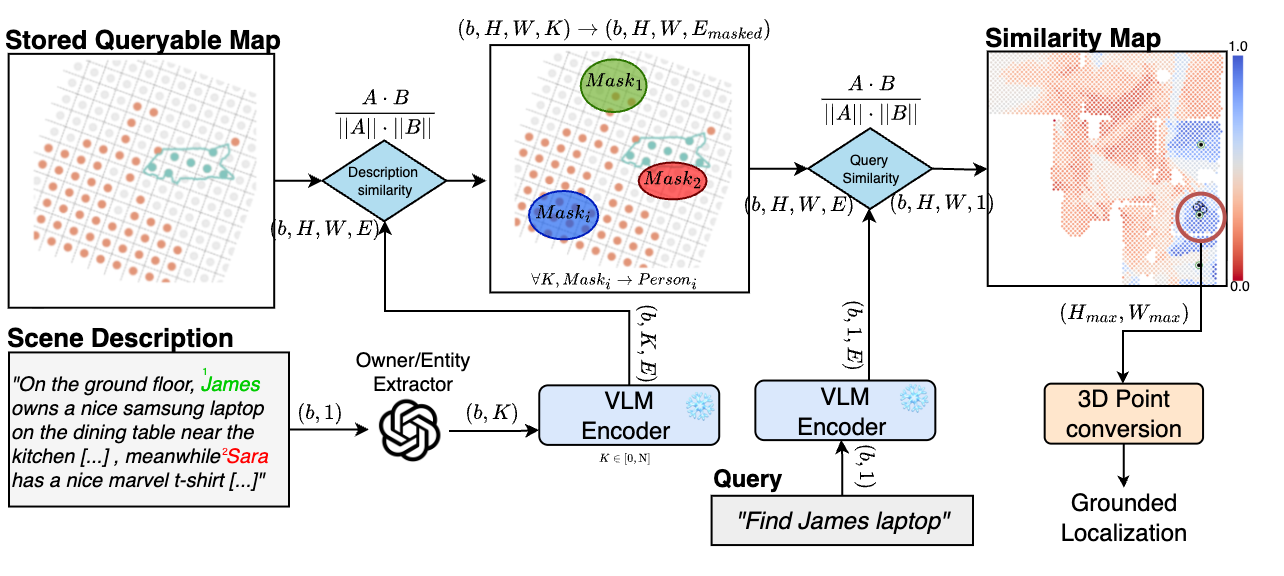}
    \caption{\textbf{POG baseline method.} We developed a ``Region-gated'' method trough the use of cosine similarity for the K descriptions, selecting the top-k regions for each K, and then another query similarity to predict the 3D point in the map.}
    \label{fig:pog_baseline}
\end{figure}

\noindent\textbf{Region-gated Personalized Grounding (RPG).} 
We developed and tested a zero-shot method for grounding textual queries in a 2D semantic feature map, as shown in Figure~\ref{fig:pog_baseline}. The input feature map is built as defined previously.
We then use an LLM (GPT-4.1 in our case) to extract K descriptive phrases from each summary, corresponding to the K people present (e.g., "Clara owns a white bed").
This ensure that the final K forms are \textit{ “person x owns object y in room z” }

Each phrase is encoded into embeddings of size $(K, E)$ using a frozen text encoder.
We then compute cosine similarity scores between these embeddings and the semantic map embeddings, selecting the top-3 matching location indices for each phrase while masking out the remaining regions, yielding a masked feature map $(H_{\text{mask}}, W_{\text{mask}}, E)$.
In addition, we preserve all pixels within a $1$-pixel radius of the selected predictions (corresponding to $1$ meter in the real world).

Next, we score the embedded query vector of size $(1,E)$ against the non-masked regions of the feature map obtained in the previous step.
From all candidate sub-regions, we choose the location with the highest similarity score as the final prediction, relying only on the pixel with the maximum similarity.
The full procedure is summarized in Algorithm~\ref{alg:simple-rgg}.

While the task formally requires outputting a single $(x,y)$ coordinate, in practice a robot must physically navigate to that location.
Therefore, we argue that top-$k$ predictions (e.g., top-3) are also meaningful: the robot can attempt to reach the first candidate, and if unsuccessful, proceed to the next predicted position.

\begin{algorithm}[!ht]
\small
\caption{Region-gated Personalized Grounding \textbf{RPG}.}
\label{alg:simple-rgg}
\begin{algorithmic}[1]
\Require $F$ (feature map, $B{\times}H{\times}W{\times}E$),
         $D$ (descriptions, $B{\times}K{\times}E$),
         $q$ (query, $B{\times}E$)
\Require $\texttt{top\_k}$, $\texttt{nhood}$, $\texttt{nms\_r}$
\Ensure $\texttt{index}\in\mathbb{N}^B$, $\texttt{value}\in\mathbb{R}^B$
\State L2-normalize $F$ and $D$ along the embedding dim.
\State $S_{\text{desc}} \gets \text{CosSim}(F, D)$ \Comment{per-pixel $\times$ per-description scores; $B{\times}H{\times}W{\times}K$}
\State $P \gets \text{NMS-TopK}(S_{\text{desc}}, \texttt{top\_k}, \texttt{nms\_r})$ \Comment{peaks per description}
\State $M \gets \text{Dilate}(P, \texttt{nhood})$ \Comment{binary mask over $H{\times}W$}
\State $\tilde F \gets F \odot M$ \Comment{apply mask (broadcast over $E$)}
\State $V \gets \text{CosSim}(\tilde F,\ \text{Broadcast}(q, H{\times}W))$ \Comment{final similarity map $B{\times}H{\times}W$}
\State $(x^\ast,y^\ast) \gets \text{ArgMax}(V)$
\State $\texttt{index} \gets y^\ast + W\cdot x^\ast$,\quad $\texttt{value} \gets V[x^\ast,y^\ast]$
\State \Return $\texttt{index} = \left(H_{max}, W_{max}\right)$
\end{algorithmic}
\end{algorithm}

\subsection{Metrics}

In the active navigation setting, we adopt standard metrics for object-driven embodied navigation, including success rate (\textbf{SR}) and success rate weighted by path length (\textbf{SPL}). 
While SR simply measures whether the agent reaches the target, 
\[
\text{SR} = \frac{1}{N} \sum_{i=1}^{N} S_i,
\]
where $S_i = 1$ if the agent successfully reaches the goal in episode $i$, and $0$ otherwise. 

SPL also accounts for efficiency by penalizing unnecessarily long trajectories,
\[
\text{SPL} = \frac{1}{N} \sum_{i=1}^{N} S_i \cdot \frac{L_i}{\max(P_i, L_i)},
\]
where $L_i$ is the shortest path length and $P_i$ the path length taken by the agent. 

This provides the most informative metric for personalization, as it reflects the ability to directly ground memorized information. 
In addition, we report the average distance to goal (\textbf{DTG}) at the episode’s end. 
In contrast to standard query grounding benchmarks, we evaluate localization accuracy directly using SR within a 1-meter threshold, rather than IoU-based metrics, due to the lack of reliable bounding-box annotations in HM3D meshes and to better reflect realistic navigation scenarios.

\subsection{Results}

Table~\ref{tab:pon_results} reports the results for active navigation (PAN). 
Human performance is substantially higher than existing navigation baselines~\cite{busch2024one, yokoyama2024vlfm}, achieving a 63\% SR, and up to 86\% when prior environment knowledge is available, as expected (line 1-2, total split). 
Among the two common navigation methods, OneMap provides the strongest baseline with a 28\% SR and a relatively high navigation efficiency of 22\% (line 3-4, total split). 
OneMap achieves higher SPL than VLFM (SR and SPL of $\sim$10\%, 6\% respectivly), indicating that its semantic feature map provides more informative guidance and shorter paths. 
Notably, it also surpasses VLFM in success rate despite using the same detector. 
Whereas VLFM relies on a PointNav policy trained for HM3D, OneMap follows shortest paths on a grid. We attribute this advantage to consensus-filtering with high-confidence semantic beliefs, in contrast to VLFM’s heuristic filtering~\cite{busch2024one}. HM3D artifacts further exacerbate false detections.

These results highlight that the benchmark is far from saturation and provides meaningful headroom for the community. 
Moreover, we argue that one of the limitations lies in the open-set object detector: while the LLM can assign the correct caption, these captions are often too long or complex to be effectively recognized by the detector.

Regarding personalized query grounding (POG), results are shown in Table~\ref{tab:pqg_results}. 
Overall, the Region-Gated method (line 5) performs best, with a SR of 25\%, as it effectively suppresses irrelevant embeddings and is therefore less sensitive to noise, but still lags well behind human performance (line 1-4). 
This gap is clearly explained by humans’ ability to interactively examine the map (e.g., rotating or zooming) before responding, which models cannot currently replicate. 
The main challenge lies in memory design: to support open-set queries, we store embeddings directly as ``memory''. 
This is particularly useful when the LLM extracts the object and its owner from the summary as descriptive information, which cannot be mapped to a fixed category.
However this comes at the cost of accuracy in grounding the correct point.
While stored graph-based memory components may help in this regard with common items, they still do not generalize to open-set personalization, leaving this as a direction for future work.

We argue that relying exclusively on large language models is not a viable solution for this benchmark. 
Continuous API calls for reasoning over memorized information are neither practical nor cost-effective, particularly in real-world household settings. 
The dataset splits are designed to highlight that simply embedding an LLM into a robot and relying on it for navigation would be prohibitively expensive and impractical.
Instead, our goal is to steer research toward lightweight or minimal LLM integration, coupled with dedicated memory mechanisms—not just databases to query, but structured feature spaces against which queries can be efficiently scored.

\setlength{\tabcolsep}{4pt}
\begin{table}[t]
\centering
\small
\renewcommand{\arraystretch}{1.12}
\begin{tabularx}{\columnwidth}{clCCCC}
\toprule
 &  & \textbf{Easy} & \textbf{Medium} & \textbf{Hard} & \textbf{Total} \\
\cmidrule(lr){3-6}
\textbf{Map} & \textbf{Method} & SR$\uparrow$ & SR$\uparrow$ & SR$\uparrow$ & SR$\uparrow$ \\
\midrule
\cellcolor{gray!10} GLB$^\dagger$ & \cellcolor{gray!10}Human 
& \cellcolor{gray!10}95.3 & \cellcolor{gray!10}89.1 & \cellcolor{gray!10}83.4 & \cellcolor{gray!10}87.0 \\
\midrule
\multirow{4}{*}{2D}
 & Random (x,y)              & 0.5   & 0.3   & 0.4   & 0.4 \\
 & Center (x,y)              & 1.2   & 0.8   & 1.0   & 1.0 \\
 & Query-Score (QS)       & 15.4  & 12.7  & 8.6   & 11.9 \\
 & Region-Gated (RPG)        & \textbf{25.3} & \textbf{23.6} & \textbf{15.2} & \textbf{16.8} \\
\bottomrule
\end{tabularx}
\caption{\textbf{POG Results.} $^\dagger$Human subjects were given the original GLB scene files.}
\label{tab:pqg_results}
\end{table}

%% file: sec/6_conclusions.tex
\section{REAL-WORLD TRANSFERABILITY}
\label{sec:discussion}

\begin{figure}[!ht]
    \centering
    \includegraphics[width=.98\columnwidth]{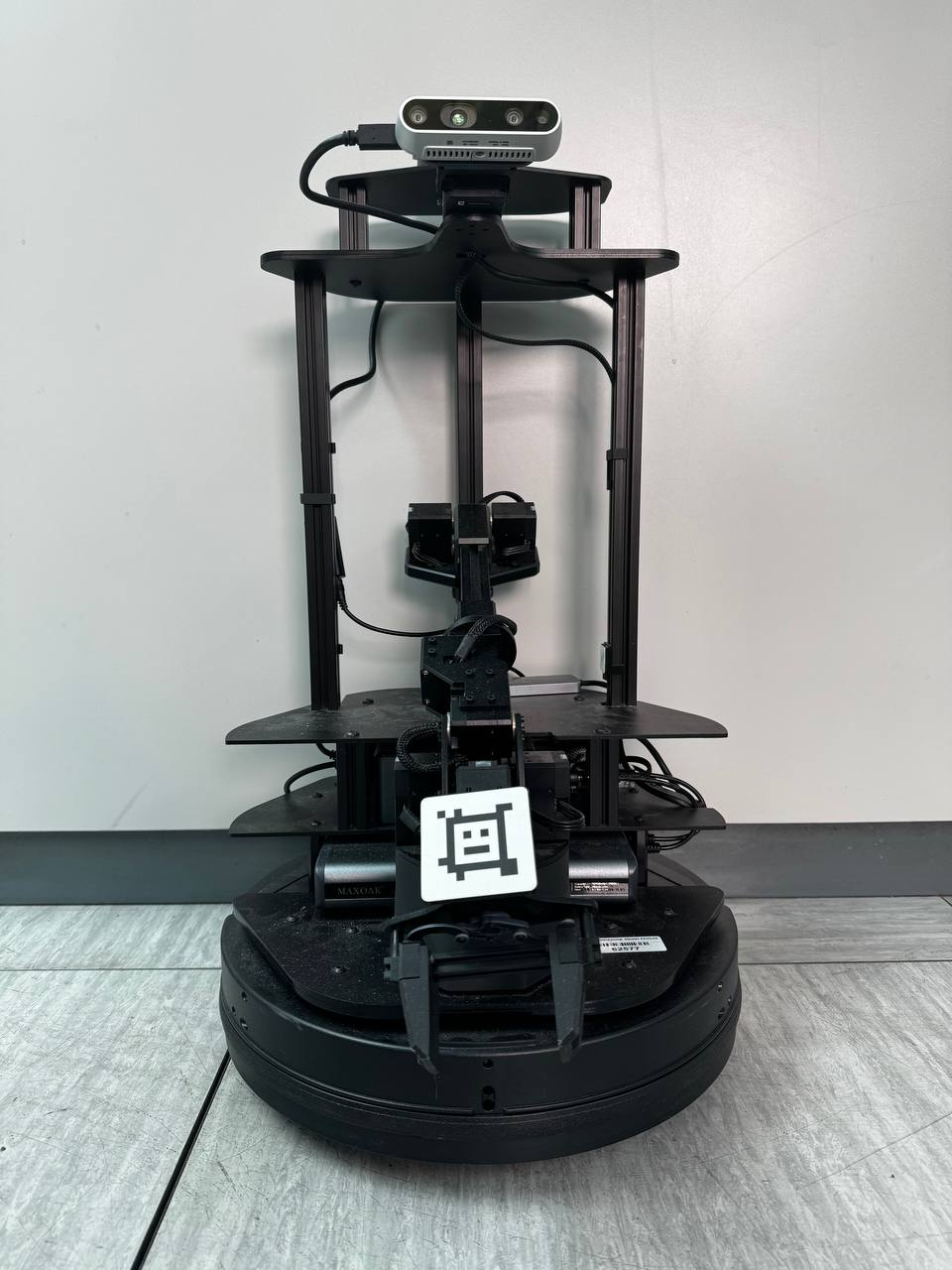}
    \caption{The LoCoBot wx250s, a low-cost mobile robot with a differential-drive base, a 6-DoF WidowX 250s arm, and an RGB-D sensor.}
    \label{fig:robot}
\end{figure}

Although \approach is entirely simulation-based within the Habitat framework~\cite{savva2019habitat}, its design targets real-world applicability. 
The use of photo-realistic HM3D scenes~\cite{ramakrishnan2021habitat}, along with common sensor configurations and action spaces compatible with common robotic platforms (e.g., Spot, TurtleBot, LocoBot), facilitates direct policy transfer. 

To demonstrate real-world feasibility, we evaluated \approach on a LoCoBot wx250s equipped with an Intel RealSense D435 camera and Hector SLAM for odometry. 
We mapped an office spanning three different rooms office environments trough a simple frontier-exploration policy until no new frontier was found (approx. 500 steps), selecting objects that were either long-tail or relatively small compared to typical office items (e.g., “art book,” “black trashcan,” “Amazon’s bottle”).
We then evaluated POG localization using the stored feature map, where our RPG algorithm successfully localized objects in 5 out of 15 cases (in accordance with simulated results). Each episode was manually annotated: we wrote three scene descriptions reflecting different difficulty levels, and for each description we defined five queries. 
The objective was to assess the effective transferability of the task to real-world scenarios, where \approach benchmark is most needed.


\section{LIMITATIONS}
\approach serves as a strong testbed for evaluating personalized Embodied AI agents, but it also highlights limitations in current navigation pipelines. 
When analyzing human performance, we observe that humans achieve consistently high results once they are familiar with the environment. 
Even on the ‘difficult’ split, their performance drops only slightly compared to the easy split. This reflects the strengths of lifelong learning, where humans excel. 
Motivated by this, we stress the need to extend more benchmarks to support lifelong settings, enabling agents to improve navigation efficiency as they acquire experience and knowledge of the environment.

Moreover, personalization is inherently dynamic: personalized information can change over short or long timeframes (e.g., ``I bought a new mug with a smiley face, store it in my cabinet''). 
Current embodied datasets do not account for this variability, and a truly personalized embodied agent must be able to operate in dynamic environments.

Finally, we argue that relying solely on large language models (LLMs) is insufficient to solve personalization, as continuous API interactions for reasoning over stored information are neither practical nor cost-effective in real-world deployments. 
More fundamentally, future robots should not separate reasoning from action—as is often done by delegating reasoning to an LLM and control to an RL model—but instead integrate them into a unified process. 
In humans, cognition and action are inherently coupled, and we believe embodied agents should follow the same principle~\cite{durante2025interactive, fung2025embodied}.

\section{CONCLUSIONS}
\label{sec:conclusions}

We introduced \approach a novel benchmark to tackle Personalization in Embodied AI encompassing both active navigation and query grounding scenarios, depending on whether agents posses prior knowledge of the environment. 
To support this benchmark, we jointly released a dataset, structured into multiple difficulty levels designed to ensure long-term relevance and challenge.
We evaluated several baseline approaches in a setting where agents receive detailed scene descriptions based on human-object ownership graphs, along with personalized queries referencing specific objects. 
Furthermore, we proposed a simple zero-shot approach that effectively addresses the grounding task in this personalized context.
Our experimental results indicate that existing approaches fall significantly short of human performance.
Overall \approach establishes a new testbed for future research on personalized embodied AI. 

Future works will target a refined \approach2.0 release, including accurate ground-truth bounding boxes to support tasks such as 3D Question Answering. 
We also plan to integrate humans into the environments, enabling dynamic multi-target queries like ``find Julia’s laptop and bring it to her'', to enhance practical applications.


%% file: root.bbl
\begin{thebibliography}{10}
\providecommand{\url}[1]{#1}
\csname url@rmstyle\endcsname
\providecommand{\newblock}{\relax}
\providecommand{\bibinfo}[2]{#2}
\providecommand\BIBentrySTDinterwordspacing{\spaceskip=0pt\relax}
\providecommand\BIBentryALTinterwordstretchfactor{4}
\providecommand\BIBentryALTinterwordspacing{\spaceskip=\fontdimen2\font plus
\BIBentryALTinterwordstretchfactor\fontdimen3\font minus \fontdimen4\font\relax}
\providecommand\BIBforeignlanguage[2]{{%
\expandafter\ifx\csname l@#1\endcsname\relax
\typeout{** WARNING: IEEEtran.bst: No hyphenation pattern has been}%
\typeout{** loaded for the language `#1'. Using the pattern for}%
\typeout{** the default language instead.}%
\else
\language=\csname l@#1\endcsname
\fi
#2}}

\bibitem{yang20253d}
Y.~Yang, H.~Yang, J.~Zhou, P.~Chen, H.~Zhang, Y.~Du, and C.~Gan, ``3d-mem: 3d scene memory for embodied exploration and reasoning,'' in \emph{Proceedings of the Computer Vision and Pattern Recognition Conference}, 2025, pp. 17\,294--17\,303.

\bibitem{liang2023code}
J.~e.~a. Liang, ``Code as policies: Language model programs for embodied control,'' in \emph{2023 IEEE Int. Conf. Robot. Autom. (ICRA)}, 2023, pp. 9493--9500.

\bibitem{song2023llm}
C.~e.~a. Song, ``Llm-planner: Few-shot grounded planning for embodied agents,'' in \emph{Proc. IEEE/CVF Int. Conf. Comput. Vis. (ICCV)}, 2023, pp. 2998--3009.

\bibitem{ziliotto2025tango}
F.~Ziliotto, T.~Campari, L.~Serafini, and L.~Ballan, ``Tango: training-free embodied ai agents for open-world tasks,'' in \emph{Proceedings of the Computer Vision and Pattern Recognition Conference}, 2025, pp. 24\,603--24\,613.

\bibitem{li2024tina}
D.~Li, W.~Chen, and X.~Lin, ``Tina: Think, interaction, and action framework for zero-shot vision language navigation,'' \emph{arXiv preprint arXiv:2403.08833}, 2024.

\bibitem{wang2024vlm}
B.~Wang, J.~Zhang, S.~Dong, I.~Fang, and C.~Feng, ``Vlm see, robot do: Human demo video to robot action plan via vision language model,'' \emph{arXiv preprint arXiv:2410.08792}, 2024.

\bibitem{kim2024survey}
Y.~Kim, D.~Kim, J.~Choi, J.~Park, N.~Oh, and D.~Park, ``A survey on integration of large language models with intelligent robots,'' \emph{Intelligent Service Robotics}, vol.~17, no.~5, pp. 1091--1107, 2024.

\bibitem{pham2025plvm}
C.~Pham, H.~Phan, D.~Doermann, and Y.~Tian, ``Plvm: A tuning-free approach for personalized large vision-language model,'' in \emph{Proceedings of the Computer Vision and Pattern Recognition Conference}, 2025, pp. 3632--3641.

\bibitem{alaluf2024myvlm}
Y.~Alaluf, E.~Richardson, S.~Tulyakov, K.~Aberman, and D.~Cohen-Or, ``Myvlm: Personalizing vlms for user-specific queries,'' in \emph{European Conference on Computer Vision}.\hskip 1em plus 0.5em minus 0.4em\relax Springer, 2024, pp. 73--91.

\bibitem{pham2024personalized}
C.~Pham, H.~Phan, D.~Doermann, and Y.~Tian, ``Personalized large vision-language models,'' \emph{arXiv preprint arXiv:2412.17610}, 2024.

\bibitem{barsellotti2024personalized}
L.~Barsellotti, R.~Bigazzi, M.~Cornia, L.~Baraldi, and R.~Cucchiara, ``Personalized instance-based navigation toward user-specific objects in realistic environments,'' \emph{Advances in Neural Information Processing Systems}, vol.~37, pp. 11\,228--11\,250, 2024.

\bibitem{dai2024think}
Y.~Dai, R.~Peng, S.~Li, and J.~Chai, ``Think, act, and ask: Open-world interactive personalized robot navigation,'' in \emph{2024 IEEE international conference on robotics and automation (ICRA)}.\hskip 1em plus 0.5em minus 0.4em\relax IEEE, 2024, pp. 3296--3303.

\bibitem{savva2019habitat}
M.~Savva, A.~Kadian, O.~Maksymets, Y.~Zhao, E.~Wijmans, B.~Jain, J.~Straub, J.~Liu, V.~Koltun, J.~Malik, \emph{et~al.}, ``{Habitat: A platform for embodied AI research},'' 2019.

\bibitem{kolve2017ai2thor}
E.~Kolve, R.~Mottaghi, R.~Han, Y.~Zhu, and A.~Gupta, ``Ai2-thor: An interactive 3d environment for visual ai,'' in \emph{arXiv preprint arXiv:1712.05474}, 2017.

\bibitem{ramakrishnan2021habitat}
S.~K. Ramakrishnan, A.~Gokaslan, E.~Wijmans, O.~Maksymets, A.~Clegg, J.~Turner, E.~Undersander, W.~Galuba, A.~Westbury, A.~X. Chang, \emph{et~al.}, ``{Habitat-matterport 3D dataset (HM3D): 1000 large-scale 3D environments for embodied AI},'' \emph{arXiv preprint arXiv:2109.08238}, 2021.

\bibitem{chang2017matterport}
A.~X. Chang, A.~Dai, T.~Funkhouser, M.~Halber, M.~Niessner, M.~Savva, S.~Song, A.~Zeng, and Y.~Zhang, ``{Matterport3D}: Learning from {RGB-D} data in indoor environments,'' 2017.

\bibitem{batra2020objectnav}
D.~Batra, A.~Gokaslan, A.~Kembhavi, O.~Maksymets, R.~Mottaghi, M.~Savva, A.~Toshev, and E.~Wijmans, ``{ObjectNav} revisited: On evaluation of embodied agents navigating to objects,'' \emph{arXiv preprint arXiv:2006.13171}, 2020.

\bibitem{krantz2022instance}
J.~Krantz, S.~Lee, J.~Malik, D.~Batra, and D.~S. Chaplot, ``Instance-specific image goal navigation: Training embodied agents to find object instances,'' \emph{arXiv preprint arXiv:2211.15876}, 2022.

\bibitem{anderson2018vision}
P.~Anderson, Q.~Wu, D.~Teney, J.~Bruce, M.~Johnson, N.~S{\"u}nderhauf, I.~Reid, S.~Gould, and A.~van~den Hengel, ``Vision-and-language navigation: Interpreting visually-grounded navigation instructions in real environments,'' 2018.

\bibitem{wani2020multion}
S.~Wani, S.~Patel, U.~Jain, A.~Chang, and M.~Savva, ``Multion: Benchmarking semantic map memory using multi-object navigation,'' \emph{Advances in Neural Information Processing Systems}, vol.~33, pp. 9700--9712, 2020.

\bibitem{chaplot2020object}
D.~S. Chaplot, D.~P. Gandhi, A.~Gupta, and R.~Salakhutdinov, ``Object goal navigation using goal-oriented semantic exploration,'' 2020.

\bibitem{krantz2023navigating}
J.~Krantz, T.~Gervet, K.~Yadav, A.~Wang, C.~Paxton, R.~Mottaghi, D.~Batra, J.~Malik, S.~Lee, and D.~S. Chaplot, ``Navigating to objects specified by images,'' 2023.

\bibitem{yokoyama2024hm3d}
N.~Yokoyama, R.~Ramrakhya, A.~Das, D.~Batra, and S.~Ha, ``Hm3d-ovon: A dataset and benchmark for open-vocabulary object goal navigation,'' \emph{arXiv preprint arXiv:2409.14296}, 2024.

\bibitem{khanna2024goat}
M.~Khanna, R.~Ramrakhya, G.~Chhablani, S.~Yenamandra, T.~Gervet, M.~Chang, Z.~Kira, D.~S. Chaplot, D.~Batra, and R.~Mottaghi, ``Goat-bench: A benchmark for multi-modal lifelong navigation,'' in \emph{Proceedings of the IEEE/CVF Conference on Computer Vision and Pattern Recognition}, 2024, pp. 16\,373--16\,383.

\bibitem{majumdar2022zson}
A.~Majumdar, G.~Aggarwal, B.~Devnani, J.~Hoffman, and D.~Batra, ``Zson: Zero-shot object-goal navigation using multimodal goal embeddings,'' \emph{Advances in Neural Information Processing Systems}, vol.~35, pp. 32\,340--32\,352, 2022.

\bibitem{yadav2025findingdory}
K.~Yadav, Y.~Ali, G.~Gupta, Y.~Gal, and Z.~Kira, ``Findingdory: A benchmark to evaluate memory in embodied agents,'' \emph{arXiv preprint arXiv:2506.15635}, 2025.

\bibitem{chang2023goat}
M.~Chang, T.~Gervet, M.~Khanna, S.~Yenamandra, D.~Shah, S.~Y. Min, K.~Shah, C.~Paxton, S.~Gupta, D.~Batra, \emph{et~al.}, ``Goat: Go to any thing,'' \emph{arXiv preprint arXiv:2311.06430}, 2023.

\bibitem{hwang2024promptable}
M.~Hwang, L.~Weihs, C.~Park, K.~Lee, A.~Kembhavi, and K.~Ehsani, ``Promptable behaviors: Personalizing multi-objective rewards from human preferences,'' in \emph{Proceedings of the IEEE/CVF Conference on Computer Vision and Pattern Recognition}, 2024, pp. 16\,216--16\,226.

\bibitem{rebuffi2017learning}
S.-A. Rebuffi, H.~Bilen, and A.~Vedaldi, ``Learning multiple visual domains with residual adapters,'' \emph{Advances in neural information processing systems}, vol.~30, 2017.

\bibitem{hu2022lora}
E.~J. Hu, Y.~Shen, P.~Wallis, Z.~Allen-Zhu, Y.~Li, S.~Wang, L.~Wang, W.~Chen, \emph{et~al.}, ``Lora: Low-rank adaptation of large language models.'' \emph{ICLR}, vol.~1, no.~2, p.~3, 2022.

\bibitem{deitke2022procthor}
M.~Deitke, E.~VanderBilt, A.~Herrasti, L.~Weihs, J.~Salvador, K.~Ehsani, W.~Han, E.~Kolve, A.~Farhadi, A.~Kembhavi, \emph{et~al.}, ``{Procthor: Large-scale Embodied AI using procedural generation},'' \emph{arXiv preprint arXiv:2206.06994}, 2022.

\bibitem{li2021ion}
W.~Li, X.~Song, Y.~Bai, S.~Zhang, and S.~Jiang, ``Ion: Instance-level object navigation,'' in \emph{Proceedings of the 29th ACM international conference on multimedia}, 2021, pp. 4343--4352.

\bibitem{gibson}
F.~Xia, A.~R. Zamir, Z.~He, A.~Sax, J.~Malik, and S.~Savarese, ``{Gibson Env: Real-World Perception for Embodied Agents},'' 2018.

\bibitem{deitke2020robothor}
M.~Deitke, W.~Han, A.~Herrasti, A.~Kembhavi, E.~Kolve, R.~Mottaghi, J.~Salvador, D.~Schwenk, E.~VanderBilt, M.~Wallingford, \emph{et~al.}, ``Robothor: An open simulation-to-real embodied ai platform,'' in \emph{Proceedings of the IEEE/CVF conference on computer vision and pattern recognition}, 2020, pp. 3164--3174.

\bibitem{busch2024one}
F.~L. Busch, T.~Homberger, J.~Ortega-Peimbert, Q.~Yang, and O.~Andersson, ``One map to find them all: Real-time open-vocabulary mapping for zero-shot multi-object navigation,'' \emph{arXiv preprint arXiv:2409.11764}, 2024.

\bibitem{minderer2024scaling}
M.~Minderer, A.~Gritsenko, and N.~Houlsby, ``Scaling open-vocabulary object detection,'' \emph{Advances in Neural Information Processing Systems}, vol.~36, 2024.

\bibitem{liu2024grounding}
S.~Liu, Z.~Zeng, T.~Ren, F.~Li, H.~Zhang, J.~Yang, Q.~Jiang, C.~Li, J.~Yang, H.~Su, \emph{et~al.}, ``Grounding dino: Marrying dino with grounded pre-training for open-set object detection,'' in \emph{European conference on computer vision}.\hskip 1em plus 0.5em minus 0.4em\relax Springer, 2024, pp. 38--55.

\bibitem{yokoyama2024vlfm}
N.~Yokoyama, S.~Ha, D.~Batra, J.~Wang, and B.~Bucher, ``Vlfm: Vision-language frontier maps for zero-shot semantic navigation,'' in \emph{2024 IEEE International Conference on Robotics and Automation (ICRA)}.\hskip 1em plus 0.5em minus 0.4em\relax IEEE, 2024, pp. 42--48.

\bibitem{yamauchi1998frontier}
B.~Yamauchi, ``Frontier-based exploration using multiple robots,'' in \emph{Proceedings of the second international conference on Autonomous agents}, 1998, pp. 47--53.

\bibitem{li2023blip}
J.~Li, D.~Li, S.~Savarese, and S.~Hoi, ``Blip-2: Bootstrapping language-image pre-training with frozen image encoders and large language models,'' in \emph{International conference on machine learning}.\hskip 1em plus 0.5em minus 0.4em\relax PMLR, 2023, pp. 19\,730--19\,742.

\bibitem{durante2025interactive}
Z.~Durante, R.~Gong, B.~Sarkar, N.~Wake, R.~Taori, P.~Tang, S.~Lakshmikanth, K.~Schulman, A.~Milstein, H.~Vo, \emph{et~al.}, ``An interactive agent foundation model,'' in \emph{Proceedings of the Computer Vision and Pattern Recognition Conference}, 2025, pp. 3652--3662.

\bibitem{fung2025embodied}
P.~Fung, Y.~Bachrach, A.~Celikyilmaz, K.~Chaudhuri, D.~Chen, W.~Chung, E.~Dupoux, H.~Gong, H.~J{\'e}gou, A.~Lazaric, \emph{et~al.}, ``Embodied ai agents: Modeling the world,'' \emph{arXiv preprint arXiv:2506.22355}, 2025.

\end{thebibliography}
